\documentclass[11pt, a4paper, logo, copyright]{eai}
\usepackage{shlab}

\usepackage[authoryear, sort&compress, round]{natbib}
\usepackage[]{mdframed}

\usepackage{dblfloatfix}

\usepackage{amsmath,amsfonts,bm}
\usepackage{multirow}
\usepackage{subcaption}
\usepackage{wrapfig}

\def\eqref#1{equation~\ref{#1}}

\def\1{\bm{1}}

\DeclareMathAlphabet{\mathsfit}{\encodingdefault}{\sfdefault}{m}{sl}
\SetMathAlphabet{\mathsfit}{bold}{\encodingdefault}{\sfdefault}{bx}{n}

\usepackage{listings}
\usepackage{hyperref}
\usepackage{url}
\usepackage{graphicx}
\usepackage{amsmath} %
\usepackage{mathrsfs} %
\usepackage{etoolbox}
\usepackage{cleveref}
\usepackage{tcolorbox}
\usepackage{colortbl}
\usepackage{booktabs}       %
\usepackage{amsfonts}       %
\usepackage{nicefrac}       %
\usepackage{microtype}      %
\usepackage{caption}
\usepackage{subcaption}
\usepackage{algorithm}
\usepackage{algorithmic}
\usepackage{multirow}
\usepackage{wrapfig}
\usepackage{bm}

\definecolor{ForestGreen}{rgb}{0.13,0.55,0.13}

\DeclareUnicodeCharacter{2705}{$\checkmark$}

\newlength\savewidth

\title{EBench: Elemental Diagnosis of Generalist Mobile Manipulation Policies}
\correspondingauthor{\fontsize{8}{10}\selectfont\textsuperscript{\textdagger}Corresponding author: Xing Gao and Hanqing Wang (\texttt{gaoxing@pjlab.org.cn;hanqingwang.c@gmail.com}).}

\author[1,2]{Ning Gao}
\author[1,3]{Jinliang Zheng}
\author[1]{Xing Gao\textsuperscript{\textdagger}}
\author[1]{Haoxiang Ma}
\author[1]{Hanqing Wang\textsuperscript{\textdagger}}
\author[1]{Yukai Wang}
\author[1]{Jiantong Chen}
\author[1]{Zanxin Chen}
\author[1,6]{Shujie Zhang}
\author[1]{Mingda Jia}
\author[1,7]{Jian Mao}
\author[1]{Xuekun Jiang}
\author[1]{Zihou Zhu}
\author[1]{Xinyu Li}
\author[1]{Shuai Wang}
\author[1,8]{Hao Li}
\author[1]{Wenzhe Cai}
\author[1]{Yuqiang Yang}
\author[1]{Xudong Xu}
\author[1]{Zhaoyang Lyu}
\author[1,4]{Yao Mu}
\author[1]{Tai Wang}
\author[1]{Jiangmiao Pang}
\author[1]{Jia Zeng}
\author[1,4]{Weinan Zhang}
\author[1,5]{Chunhua Shen}

\affil[1]{Shanghai AI Laboratory}
\affil[2]{Xi'an Jiaotong University}
\affil[3]{Institute for AI Industry Research (AIR), Tsinghua University}
\affil[4]{Shanghai Jiao Tong University}
\affil[5]{Zhejiang University}
\affil[6]{Tsinghua University}
\affil[7]{Tongji University}
\affil[8]{University of Science and Technology of China}

\begin{document}

\begin{abstract}
\links{
  \link{homepage}{Project Page}{https://internrobotics.github.io/EBench-home/},
  \link{github}{Code}{https://github.com/InternRobotics/EBench},
  \link{leaderboard}{Leaderboard}{https://internrobotics.shlab.org.cn/eval/evaluation-list}
}
\vspace{0.5em}

We present EBench, a simulation benchmark that diagnoses generalist mobile manipulation policies beyond a single success-rate scalar.
EBench comprises 26 diverse and challenging manipulation tasks annotated along 5 capability dimensions and 4 generalization dimensions.
We evaluate state-of-the-art generalist manipulation models including $\pi_0$, $\pi_{0.5}$, XVLA, and InternVLA-A1, and reveal that the models exhibit strikingly different capability profiles:
$\pi_{0.5}$ achieves the highest test success rate, the best train--test retention, and the strongest mobile manipulation performance; $\pi_0$ leads on dexterous fixed-base and high-precision tasks; XVLA and InternVLA-A1 exhibit complementary strengths across atomic skills and operating regimes.
Beyond capability profiling, EBench analyzes the generalization ability from 4 representative perspectives, identifying the impact of different distribution shift factors. The results reveal strengths and weaknesses of models behind an overall score. We hope this benchmark offers a broad set of diagnostic signals to guide iteration on generalist manipulation models.

\end{abstract}

\maketitle

\begin{figure*}[h]
\centering
\includegraphics[width=\linewidth]{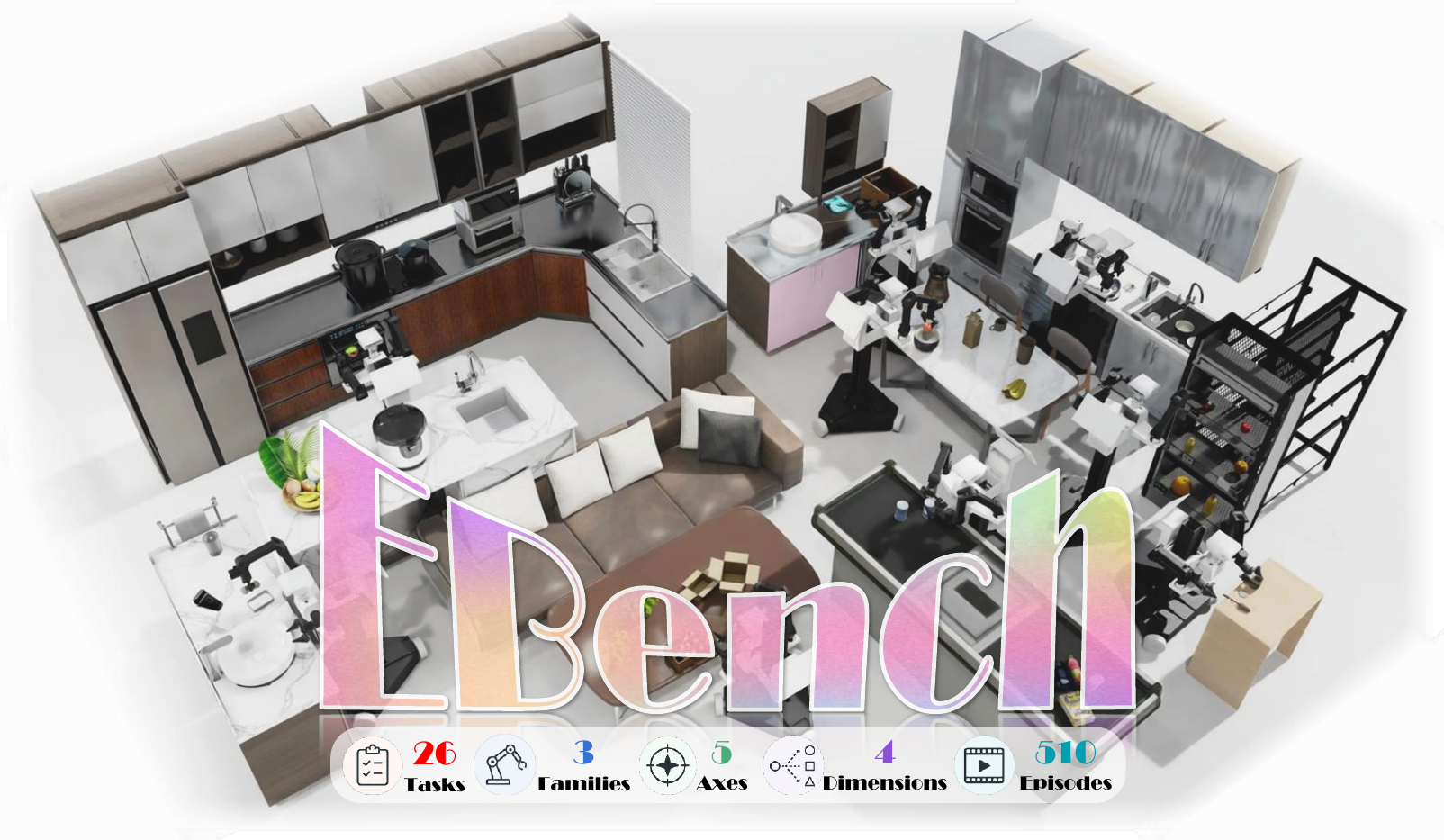}
\caption{\textbf{EBench} is a simulation benchmark for generalist embodied manipulation that, within a single evaluation suite, simultaneously covers long-horizon, dexterous-and-precise, and mobile manipulation across 9 scene categories. Each of the 26 tasks is tagged along 5 capability axes and paired with 4 controlled generalization dimensions, so that a single scalar success rate decomposes into an interpretable capability profile.}
\label{fig:teaser}
\vspace{-10pt}
\end{figure*}

\section{Introduction}
\label{sec:introduction}

Despite substantial progress of simulation benchmarks, thoroughly evaluating general-purpose manipulation policies remains challenging.
State-of-the-art generalist manipulation policies now report success rates on contemporary simulation suites to demonstrate their superior performance. However, there are fundamental questions that aggregate numbers cannot answer: \emph{Where is a model strong, where does it break}, and \emph{how does that pattern shift as the deployment distribution drifts away from the training distribution?}

The gap is structural.
Single-scene tabletop suites such as RLBench~\citep{james2020rlbench}, CALVIN~\citep{mees2022calvin}, and LIBERO~\citep{liu2023libero} cover a narrow slice of physical interaction.
Larger-scale efforts such as RoboCasa~\citep{nasiriany2024robocasa}, RoboTwin~\citep{mu2025robotwin,chen2025robotwin}, and GenManip~\citep{gao2025genmanip} broaden task and embodiment coverage, but each focuses on a single regime: tabletop pick-and-place, mobile rearrangement, or one-shot manipulation. Diagnostic benchmarks such as RMBench~\citep{chen2026rmbench} isolate a single capability axis, namely memory. BEHAVIOR-1K~\citep{li2023behavior} illustrates broader task types, while falls back to overall scores.
The community lacks a benchmark whose task suite is broad enough to cover long-horizon, dexterous, and mobile regimes together, and well-instrumented to support fine-grained analysis rather than a single leaderboard scalar. Motivated by this urgency, we introduce \textbf{EBench}, a simulation benchmark for generalist manipulation that addresses these three needs together. EBench has three core contributions:
\begin{enumerate}[leftmargin=14pt, labelindent=0pt]
    \item \textbf{A benchmark codebase for mobile manipulation.} EBench's open-source infrastructure bundles three pieces normally maintained in isolation: a two-stream \emph{data-synthesis} pipeline that combines human teleoperation for dexterous-and-precise tasks with a key-frame-pose plus cuRobo~\citep{sundaralingam2023curobo} motion planner for mobile and long-horizon tasks; a composable \emph{scoring library} that assembles per-task success and partial-progress metrics from a shared set of evaluation primitives, including scene-graph relations between objects, articulation joint angles, object tilt and orientation, and temporal-ordering constraints over sub-goals; and a distributed \emph{evaluation runner} that completes the full validation split on 8 consumer GPUs in roughly 30 minutes.
    \item \textbf{Wide-spectrum tasks with rich annotations.} On top of this codebase we assemble 26 tasks that span three families rarely co-exist in a single suite: 10 mobile pick-and-place tasks, 9 long-horizon multi-stage tasks, and 7 dexterous-and-precise tasks with sub-centimetre tolerance. Scene assets are sourced from GRUtopia~\citep{wang2024grutopia} and InternScenes~\citep{zhong2026internscenes} and object assets from Objaverse~\citep{deitke2023objaverse}; Each task is then manually annotated along five dimensions: scene type, atomic skill, temporal horizon, precision, and operating mode. Aggregate scores thus decompose into interpretable capability coordinates. 
    \item \textbf{Controlled out-of-distribution evaluation via asset partitioning.} Beyond isolated train, validation, and test splits, EBench evaluates four axes: unseen backgrounds, unseen objects, paraphrased instructions, and their mixture. Train and test sets are isolated at the asset level.
\end{enumerate}

We evaluate four recent VLAs on EBench: $\pi_0$~\citep{black2024pi_0}, $\pi_{0.5}$~\citep{intelligence2025pi_}, XVLA~\citep{zheng2025x}, and InternVLA-A1~\citep{cai2026internvla}. Their aggregate test success rates range from $24.7$ to $41.0\%$, and the five-dimensional capability profiles diverge by tens of points. $\pi_{0.5}$ attains the highest test SR of $41.0\%$ and the strongest train--test retention, with a retention ratio of $0.91$. InternVLA-A1 has the biggest gap of $31$ points between mobile and dexterous fixed-base tasks. Performance rankings vary markedly across atomic skills, indicating complementary strengths across policies rather than a single uniformly dominant model. We further analyzes the generalization ability from 4 representative perspectives, identifying the impact of different distribution shift factors. The results reveal strengths and weaknesses of models behind an overall score. We hope this benchmark offers a broad set of diagnostic signals to guide iteration on generalist manipulation models.


\section{Related Work}
\label{sec:related_work}

\paragraph{Simulation benchmarks for manipulation.}
 Eval suites of tabletop tasks such as RLBench~\citep{james2020rlbench}, CALVIN~\citep{mees2022calvin}, and LIBERO~\citep{liu2023libero} pioneered standardized evaluation but cover a narrow regime of short-horizon, fixed-base pick-and-place.
Mobile and multi-scene suites such as Habitat~\citep{savva2019habitat}, SAPIEN~\citep{xiang2020sapien}, ManiSkill~\citep{mu2021maniskill}, and RoboCasa~\citep{nasiriany2024robocasa} broaden scene diversity but seldom include sub-centimetre dexterous behaviors.
Procedurally generated suites such as RoboTwin~\citep{mu2025robotwin,chen2025robotwin} and GenManip~\citep{gao2025genmanip} scale task counts but still report a per-task scalar success rate without a structured taxonomy.
Real-to-sim transfer benchmarks such as SimplerEnv~\citep{li2024evaluating} mirror a fixed set of real-robot tabletop tasks in simulation, optimising for fidelity to one embodiment rather than coverage across regimes.
Targeted diagnostic benchmarks such as RMBench~\citep{chen2026rmbench} probe a single capability axis in isolation.
EBench differs from this prior work on two axes: it hosts long-horizon, dexterous, and mobile manipulation under one evaluation protocol, and it pairs every task with a 5 capability axes and 4 generalization dimensions, so that aggregate scores decompose into interpretable coordinates rather than collapsing into a leaderboard scalar.

\paragraph{Vision--language--action models.}
$\pi_0$~\citep{black2024pi_0} and its successor $\pi_{0.5}$~\citep{intelligence2025pi_} use flow-matching action heads on top of large multi-robot pre-training mixtures.
XVLA~\citep{zheng2025x} decouples vision--language understanding from action execution via a modular decoder.
InternVLA-A1~\citep{cai2026internvla} pairs strong visual representations with a hierarchical planner.
Adjacent generalist policies include OpenVLA~\citep{kim2024openvla}, GR00T-N1~\citep{bjorck2025gr00t}, RDT~\citep{liu2025rdt}, Octo~\citep{team2024octo}, and Diffusion Policy~\citep{chi2025diffusion}; many of these are pre-trained on cross-embodiment datasets such as Open X-Embodiment~\citep{o2024open}.
Recent open codebases such as StarVLA~\citep{community2026starvla} and StarVLA-$\alpha$~\citep{ye2026starvla} explore lighter-weight and more modular VLA recipes.
These models are typically compared on hardware-specific real-robot runs or on narrow simulation subsets; EBench provides a multi-dimensional comparison of recent VLAs under a matched generalist protocol.


\section{The EBench Benchmark}
\label{sec:benchmark}

\begin{figure*}[t]
\centering
\includegraphics[width=\linewidth]{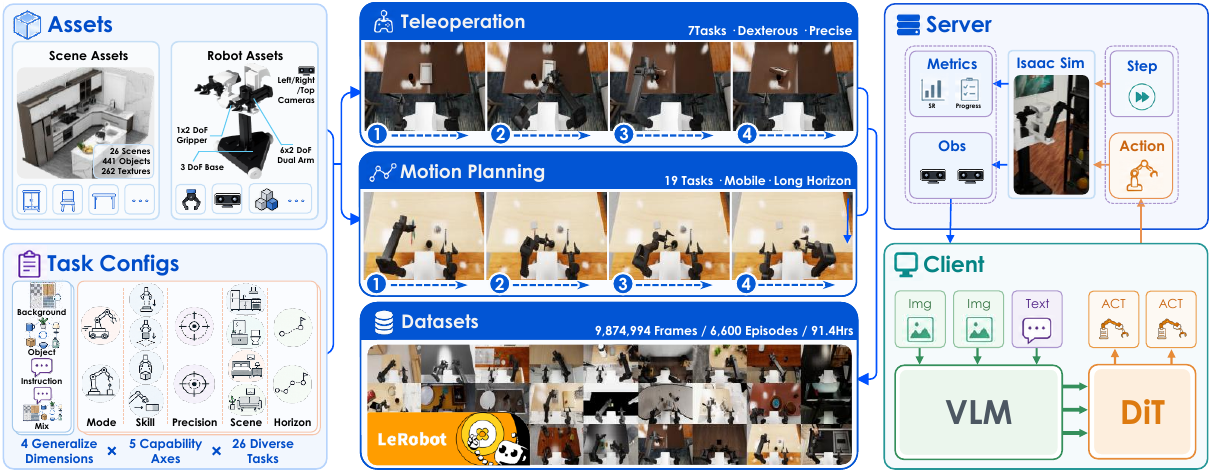}
\caption{\textbf{EBench end-to-end pipeline.}
\emph{Left}: 26 tasks span pick-and-place, long-horizon, and dexterous-and-precise families, instantiated on shared scene and robot assets.
\emph{Middle, two-track synthesis}: dexterous-and-precise demonstrations are collected via human \textbf{teleoperation} (top); mobile and long-horizon trajectories are generated by \textbf{motion planning} from key-frame end-effector poses fed to cuRobo (bottom).
\emph{Right}: EBench is evaluated through a client--server protocol. The IsaacSim-backed server returns observations and a step signal, and VLA or WAM clients (e.g.\ a VLM with a DiT action head) emit actions in response.}
\label{fig:method}
\vspace{-15pt}
\end{figure*}

\subsection{Task Design and Diversity}
\label{subsec:task_design}

EBench comprises 26 manipulation tasks organised into three families: \textbf{Mobile Pick-and-Place} (10 mobile tasks, $600$--$1{,}000$ simulation steps per episode), \textbf{Mobile Long-Horizon} (9 multi-stage mobile sequences, $3{,}000$--$5{,}000$ steps), and \textbf{Table-Top Dexterous-and-Precise} (7 fixed-base tasks, $1{,}500$--$3{,}500$ steps, covering sub-centimetre insertion, alignment, and bimanual coordination). Physics is simulated at $60$~Hz, so the longest task corresponds to roughly $83$~seconds of robot operation. Each task is annotated along five dimensions: scene, atomic skill, temporal horizon, precision, and operating mode. Table~\ref{tab:taxonomy} summarises the categories within each dimension. The taxonomy supports interpretable queries such as ``how does model $X$ perform on high-precision, long-horizon mobile tasks?'' and prevents covering up of weaknesses by strong performance on easy majority categories.

\vspace{-5pt}
\begin{table}[h]
\centering
\caption{EBench five axes task taxonomy. Numbers in parentheses indicate the number of categories within each dimension.}
\vspace{5pt}
\label{tab:taxonomy}
\begin{tabular}{@{}ll@{}}
\toprule
\textbf{Axes} & \textbf{Categories} \\
\midrule
Scene (9) & Bedroom, Bathroom, Kitchen, Living Room, Study, \\
          & Dining Room, Supermarket, Industrial, Logistics \\
Atomic Skill (11) & Grasp, Place, Push, Pull, Press, Insert, Pour, Flip, Sweep, Handover \\
Range (2) & Short Horizon ($<2,000$ steps), Long Horizon ($\geq 2,000$ steps) \\
Precision (3) & Low ($\geq 10$ cm), Medium ($<10$ cm, and $\geq 1$ cm), High ($<1$cm) \\
Operating Mode (2) & Mobile, Fixed but Dexterous  \\
\bottomrule
\end{tabular}
\end{table}

All tasks share a unified action space for a dual-arm robot mounted on a mobile base: each arm can be commanded in either 6-DoF joint position or 6-DoF end-effector pose, paired with a per-arm gripper width, and the base accepts a $3$-D velocity command (planar $x$, $y$, and yaw rate), which the model is free to emit on every task. A single model checkpoint can therefore be evaluated across regimes without any architectural modification. 

\subsection{Data Synthesis: Teleoperation and Motion Planning}
\label{subsec:data_synthesis}
To incorporate such diverse behaviors, the collection of post-training demonstration faces the following challenges:
(1) Dexterous-and-precise tasks require complex interactions that motion planner can hardly program.
(2) Collecting a successful Long-horizon demonstration is extremely hard through human teleoperation due to the exponentially amplified failure probability in the long sequence.
(3) Mobile manipulation is hard to teleoperate since a single operator has to coordinate base motion and arm motion through the same controller, and small base disturbances destabilize the arm reference frame.
To solve the challenges, EBench couples two complementary collecting streams, as shown in Figure~\ref{fig:method}:

\begin{enumerate}[leftmargin=14pt, labelindent=0pt]
\item \textbf{Teleoperation for dexterous-and-precise tasks.} The 7 dexterous tasks are collected through a kinematically isomorphic actor-follower setup. This preserves the reactive feedback and dynamic adjustments needed for contact-rich micro-corrections such as peg-in-hole insertion, nut tightening, and gear meshing. 
\item \textbf{Key-frame pose and cuRobo for mobile and long-horizon tasks.} For the remaining 19 tasks, where teleoperation is either too expensive (long-horizon) or too awkward to control (mobile), the annotator instead specifies a sparse sequence of key-frame end-effector poses, together with base waypoints for mobile cases. cuRobo~\citep{sundaralingam2023curobo} then solves a collision-free, minimum-jerk trajectory that connects them. This stream produces thousands of episodes per task without sacrificing kinematic feasibility, and the resulting trajectories are immediately re-rendered under randomized backgrounds, objects, and lighting to produce the generalization variants (\S\ref{subsec:generalization}).
\end{enumerate}
The post-training dataset contains $91.4$ hours demonstrations, $6,600$ episodes ultimately, organized in LeRobot format. Each dexterous-and-precise task contributes $400$ teleoperated episodes, each mobile pick-and-place task contributes $200$ motion-planned episodes, and each long-horizon task contributes $200$ motion-planned episodes.

\subsection{Generalization Dimensions}
\label{subsec:generalization}

EBench controls 4 generalization dimensions in evaluation:
(1) \textbf{Background} replaces scene textures and lighting with unseen variants while objects and instructions are held fixed.
(2) \textbf{Object} swaps each manipulated entity for a geometrically distinct unseen instance within the same category.
(3) \textbf{Instruction} paraphrases each natural-language command while preserving its operational goal.
(4) \textbf{Mix} applies background, object, and instruction perturbations jointly.
Background and instruction probe perceptual and linguistic robustness without changing the underlying physics, object swaps require physical generalisation, and Mix compounds the two.
Train and test sets share the same synthesis pipeline and the same randomisation axes; the training set is itself drawn under the full background, object, and instruction randomisation. Train/test isolation is enforced \emph{at the asset level}: scene textures, object instances, and instruction paraphrases used for training are drawn from a pool that is disjoint from the test pool. 

\subsection{Evaluation Protocol and Metric Library}
\label{subsec:protocol}

Since EBench is born to evaluate generalist policies, the primary protocol requires \textbf{one model checkpoint to solve all 26 tasks}.
The validation set is split into \textsc{Val-Train}, with $130$ in-distribution episodes ($5$ per task across $26$ tasks), and \textsc{Val-Unseen}, with $154$ episodes containing object swaps drawn from the unseen asset pool.
The \textsc{Test} split comprises $510$ episodes spanning all four generalisation dimensions ($20$ per task for $24$ tasks; $15$ per task for two long-horizon tasks). It is released publicly together with the rest of the benchmark, but is constructed from a pool of scene, object, and instruction assets that is disjoint from the training pool.

Two metrics are adopted: a binary success signal \textbf{SR (Primary)} and a task \textbf{Score} that rewards partial progress through the task.
The score is computed stage by stage rather than from a single distance function. Each task is declared as an ordered sequence of stages, every stage holds a set of conditions over simulator state, and the score advances whenever the next stage in sequence is satisfied; SR fires only when the final stage is reached. Stage conditions are represented by a shared schema, namely \textbf{evaluation primitives}. They are generated directly from simulator state: scene-graph relations between objects (e.g.\ ``cup on tray''), articulation joint angles for doors, drawers, and tools, object tilt and orientation, and end-effector or base pose tolerances. Composing these primitives into an ordered stage graph replaces the per-task hand-coded evaluators used in prior benchmarks and makes scores directly comparable across tasks within a family.


\section{Experimental Setup}
\label{sec:experiments}

\paragraph{Evaluated Models.}
\label{subsec:models}
We evaluate 4 recent vision--language--action (VLA) models that span distinct architecture and pre-training mixtures: $\bm{\pi_0}$~\cite{black2024pi_0}, $\bm{\pi_{0.5}}$~\cite{intelligence2025pi_}, \textbf{XVLA}~\cite{zheng2025x}, and \textbf{InternVLA-A1}~\cite{cai2026internvla}. All models are fine-tuned from pretrained checkpoints on the same EBench training data using a consistent recipe: $200$K gradient steps, batch size $128$, AdamW optimizer, and a cosine learning-rate scheduler with warm-up.

\paragraph{Post-Training and Evaluation Protocol.}
\label{subsec:details}
Post-training uses all $6{,}600$ demonstration episodes described in \S\ref{subsec:data_synthesis}, with teleoperated and motion-planned trajectories roughly balanced at the frame level.
Observations consist of RGB images at $224 \times 224$ from three viewpoints: left, right, and topdown views, together with proprioceptive state and a natural-language instruction. Each frame is recorded at the simulation rate ($60$~Hz physics step), and the policy is queried at the same rate.
Because the IsaacSim renderer is non-deterministic, each model is evaluated three times and we report the mean and standard deviation across runs.


\section{Capability Profiling}
\label{sec:results_profiling}

\begin{table*}[t]
\centering
\small
\caption{\textbf{Overall performance of baselines across evaluation splits.} \textbf{SR} (\%) as the primary metric is highlighted in \textbf{bold}; \textbf{Score} denotes continuous task progress (\%), and \textbf{Retention} is the ratio of \textsc{Test} to \textsc{Val-Train}. Results are reported as mean $\pm$ standard deviation over three evaluation runs, with the best value in each column shown in bold.
}
\label{tab:topline}
\newcommand{\pmsd}[1]{\textcolor{gray!60}{\scriptsize\sffamily\ensuremath{\pm#1}}}
\resizebox{\textwidth}{!}{%
\begin{tabular}{@{}l cc cc cc cc@{}}
\toprule
\multirow{2}{*}{\textbf{Model}}& \multicolumn{2}{c}{\textbf{\textsc{Val-Train}}} & \multicolumn{2}{c}{\textbf{\textsc{Val-Unseen}}} & \multicolumn{2}{c}{\textbf{\textsc{Test}}} & \multicolumn{2}{c}{\textbf{Retention}} \\
\cmidrule(lr){2-3} \cmidrule(lr){4-5} \cmidrule(lr){6-7} \cmidrule(lr){8-9}
 & \textbf{SR} & Score & \textbf{SR} & Score & \textbf{SR} & Score & \textbf{SR} & Score \\
\midrule
XVLA~\cite{zheng2025x} & 28.3\pmsd{2.5} & 42.1\pmsd{1.3} & 22.7\pmsd{4.3} & 35.9\pmsd{3.7} & 24.7\pmsd{1.1} & 37.5\pmsd{0.9} & 0.87 & 0.89 \\
InternVLA-A1~\cite{cai2026internvla} & 33.1\pmsd{2.0} & 44.2\pmsd{1.8} & 20.8\pmsd{1.1} & 33.8\pmsd{0.8} & 27.6\pmsd{1.6} & 40.2\pmsd{2.0} & 0.83 & 0.91 \\
$\pi_0$~\cite{black2024pi_0} & 41.3\pmsd{0.4} & 57.1\pmsd{1.2} & 32.3\pmsd{2.4} & 48.9\pmsd{0.9} & 34.4\pmsd{0.8} & 50.9\pmsd{0.9} & 0.83 & 0.89 \\
$\pi_{0.5}$~\cite{intelligence2025pi_} & \textbf{44.9\pmsd{4.2}} & \textbf{58.7\pmsd{3.3}} & \textbf{39.8\pmsd{0.7}} & \textbf{55.4\pmsd{2.9}} & \textbf{41.0\pmsd{1.2}} & \textbf{56.8\pmsd{0.8}} & \textbf{0.91} & \textbf{0.97}\\
\bottomrule
\end{tabular}%
}
\end{table*}

\subsection{Overall Performance}
\label{subsec:topline}

Table~\ref{tab:topline} shows that the four models span a sixteen-point range in test SR ($24.7$--$41.0\%$) and exhibit markedly different in-distribution behaviors. $\pi_{0.5}$ achieves the highest test SR ($41.0\%$) and the strongest retention (SR: $0.91$, Score: $0.97$), indicating that its validation performance is the most reliable predictor of held-out capability. In contrast, InternVLA-A1 drops from $33.1\%$ SR on \textsc{Val-Train} to $20.8\%$ on \textsc{Val-Unseen} and yields relatively weak retention on the held-out test split ($0.83$ SR retention), suggesting limited robustness to distribution shifts. Similarly, $\pi_0$ achieves a test SR of $34.4\%$ with the same $0.83$ SR retention.

\subsection{Five-Dimensional Capability Breakdown}
\label{subsec:five_dim_breakdown}

\begin{figure*}[t]
    \centering
    \includegraphics[width=0.96\linewidth]{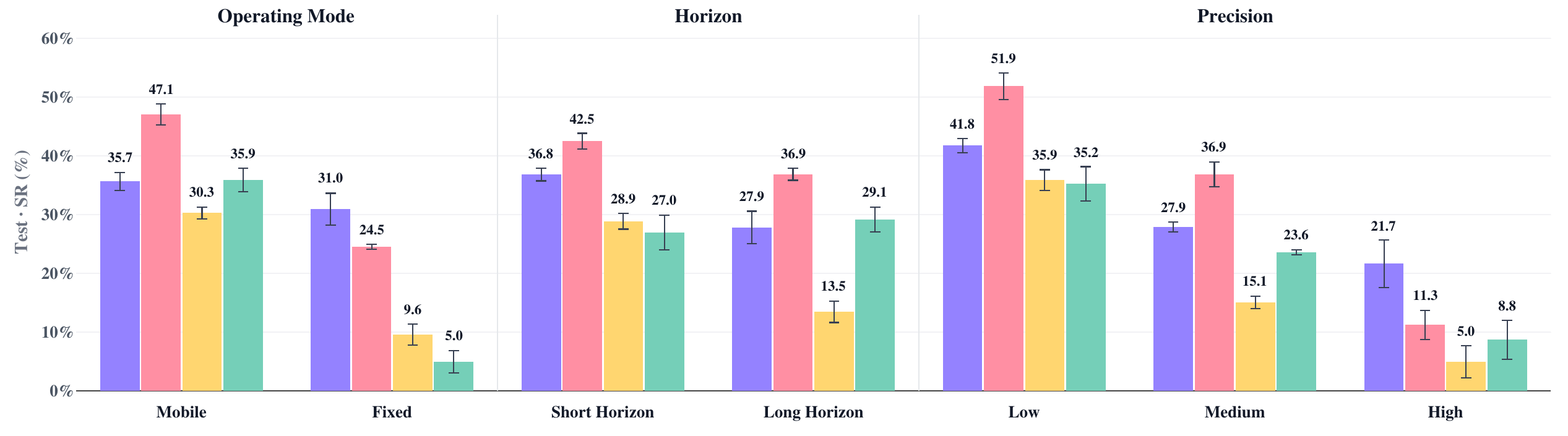}\\[0.4em]
    \includegraphics[width=0.96\linewidth]{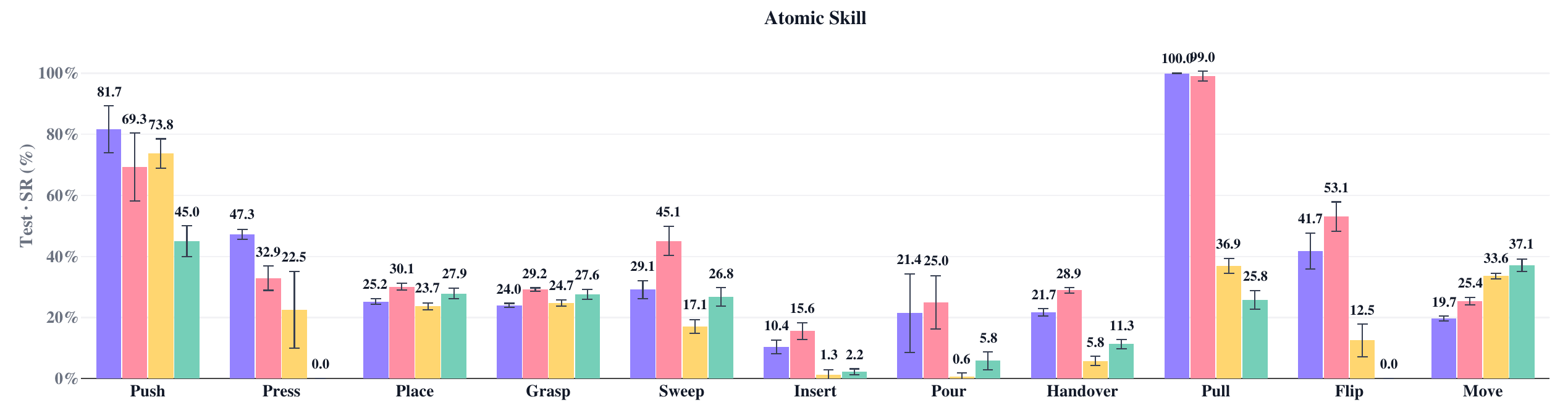}\\[0.4em]
    \includegraphics[width=0.96\linewidth]{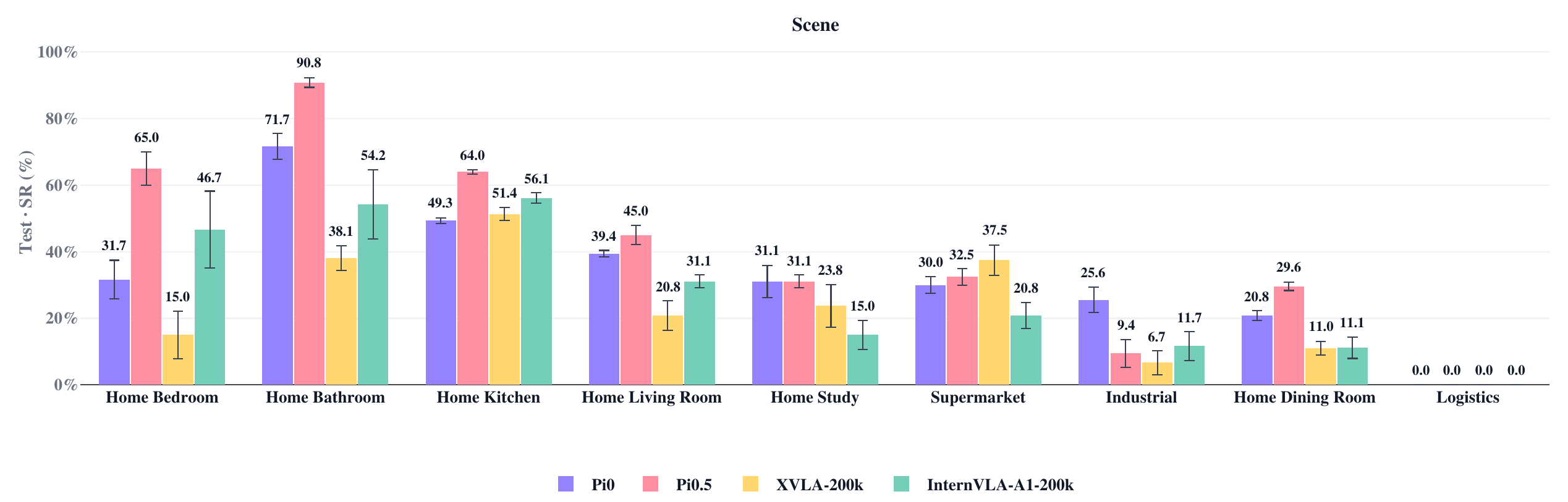}
\caption{\textbf{Capability breakdown on the five axes.} The top row reports overall success rate and three task-level axes: operating mode, temporal horizon, and precision tolerance. The middle row breaks performance down by atomic skill, while the bottom row reports performance across scene categories. Bars denote the mean test SR, and error bars denote standard deviation across seeds. 
}\label{fig:compact}
\end{figure*}

Figure~\ref{fig:compact} decomposes test SR along five complementary axes.
The top row summarizes overall performance and three low-cardinality factors, namely operating mode, temporal horizon, and precision. The middle row breaks down performance by atomic skill, and the bottom row reports scene-wise success rates. While the models differ in aggregate SR, their capability profiles also differ markedly across these dimensions.

\paragraph{Operating mode.}
InternVLA-A1 performs competitively on mobile manipulation, achieving a test SR of $35.9\%$, but its performance drops sharply on dexterous fixed-base tasks ($5.0\%$ SR). This results in the largest mobile-to-dexterous gap in the cohort ($31$ points), suggesting that the model handles navigation-scale decision making effectively but lacks the fine-grained contact control required for dexterous manipulation. In contrast, $\pi_0$ exhibits the most balanced performance profile, with a smaller $5$-point gap between mobile ($35.7\%$) and dexterous ($31.0\%$) settings, while $\pi_{0.5}$ achieves the highest mobile SR ($47.1\%$).

\paragraph{Precision and horizon.}
On sub-centimetre high-precision tasks, $\pi_0$ leads at $21.7\%$ SR, while the other models score between $5.0$ and $11.3\%$ SR.
On low-precision tasks, $\pi_{0.5}$ achieves the best performance with $51.9\%$ SR, and other models score between $35$ and $42\%$ SR.
Task horizon reveals a different pattern. Short-horizon tasks are generally easier, with all models achieving SRs in the $27$--$43\%$ range. Long-horizon tasks expose substantially larger performance gaps: $\pi_{0.5}$ achieves the highest SR ($36.9\%$), whereas XVLA drops sharply from $28.9\%$ on short tasks to $13.5\%$ on long-horizon settings, suggesting weaker temporal credit assignment in its modular decoder architecture.

\paragraph{Atomic skills and scenes.}
No single model dominates all eleven atomic skills.
$\pi_0$ leads on Pull at $100.0\%$, Push $81.7\%$,  and  Press at $47.3\%$.
XVLA reaches $73.8\%$ on Push but bottoms out on Pour at $0.6\%$.
InternVLA-A1 wins on Move yet scores $0\%$ on Press and Flip.
In contrast, InternVLA-A1 is the only model with catastrophic-zero categories. Scene-level rankings exhibit similarly heterogeneous patterns. $\pi_{0.5}$ performs best in Bedroom, Bathroom, Kitchen, Living Room, and Dining scenes, XVLA achieves the highest SR in Supermarket, and $\pi_0$ leads in Industrial settings and ties $\pi_{0.5}$ in Study.


\section{Generalization Diagnosis}
\label{sec:results_generalization}

\begin{wrapfigure}{r}{0.55\textwidth}
\includegraphics[width=0.50\textwidth]{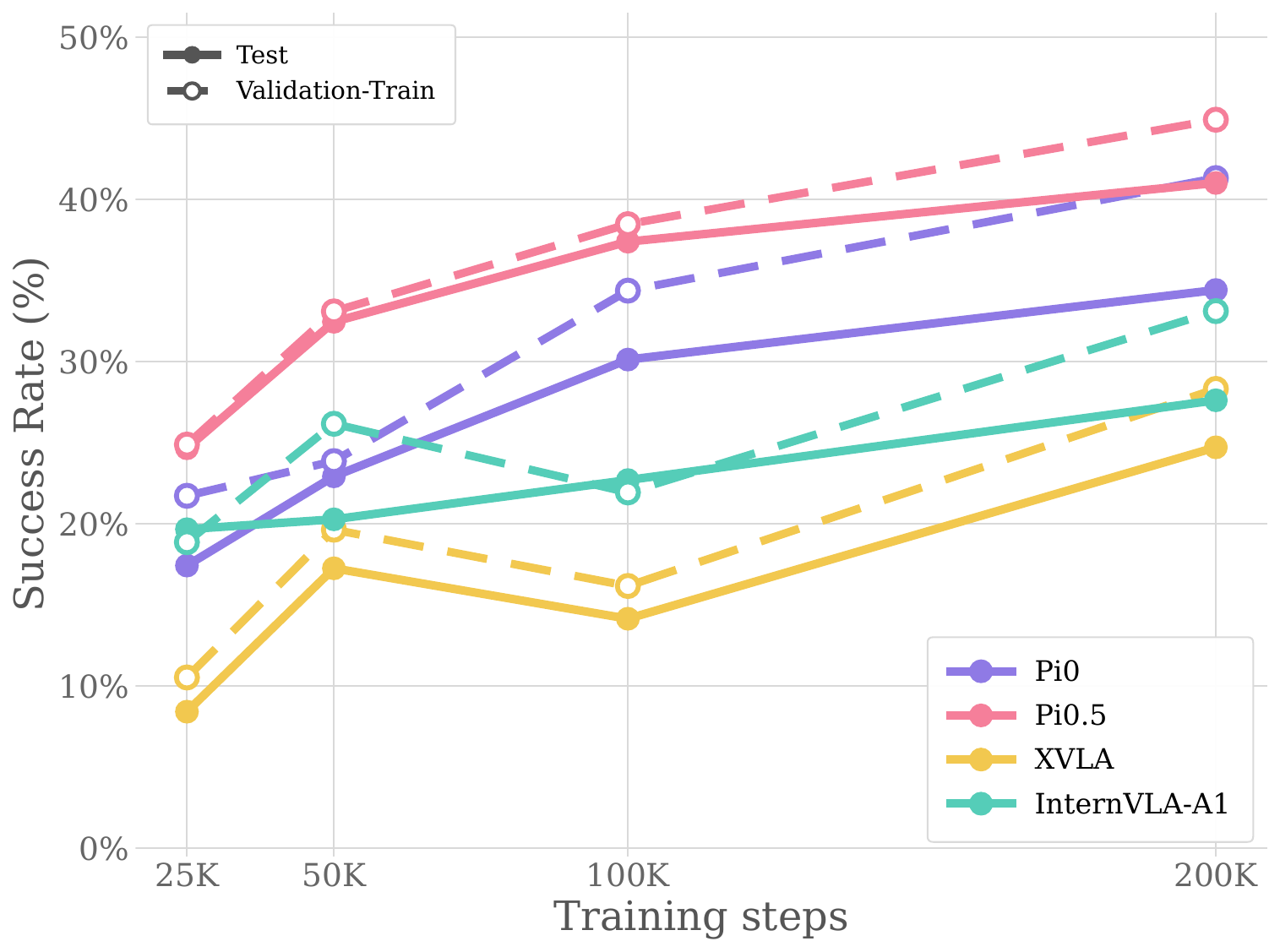}
\caption{\textbf{SR} of baselines on \textsc{Validation-Train} and \textsc{Test} split across different training steps. Dashed and solid lines denote Train and Test results, respectively.}
\label{fig:fit_generalize}
\vspace{-1.5em}
\end{wrapfigure}

Aggregate SR at a single checkpoint provide only a static view of performance.
To examine how generalization evolves during post-training, we evaluate each model at $25$k, $50$k, $100$k, and $200$k steps, and plot Validation-Train and Test SR in Figure~\ref{fig:fit_generalize}.

\paragraph{Fit--generalization dynamics.} The vertical gap between the dashed and solid curves in Figure~\ref{fig:fit_generalize}  measures the empirical fit--generalization gap: smaller gaps indicate better transfer of in-distribution gains to held-out rollouts.
Overall, additional post-training improves Test SR for all models by $200$k steps, but the transfer from Validation-Train to Test is model-dependent. Specifically,
$\pi_{0.5}$ shows the most stable dynamics, with the two curves rising largely together and the highest final Test SR.
$\pi_0$ also improves steadily, though its late-stage gap becomes more visible, indicating incomplete transfer of additional fit.
XVLA is more sensitive to training duration, showing a non-monotonic Test trajectory before recovering at $200$k.
InternVLA-A1, in contrast, shows steady improvement on Test despite a more fluctuating Validation-Train trajectory, resulting in a less consistent fit–generalization gap across training stages.

\begin{wrapfigure}{r}{0.52\textwidth}
    \includegraphics[width=0.52\textwidth]{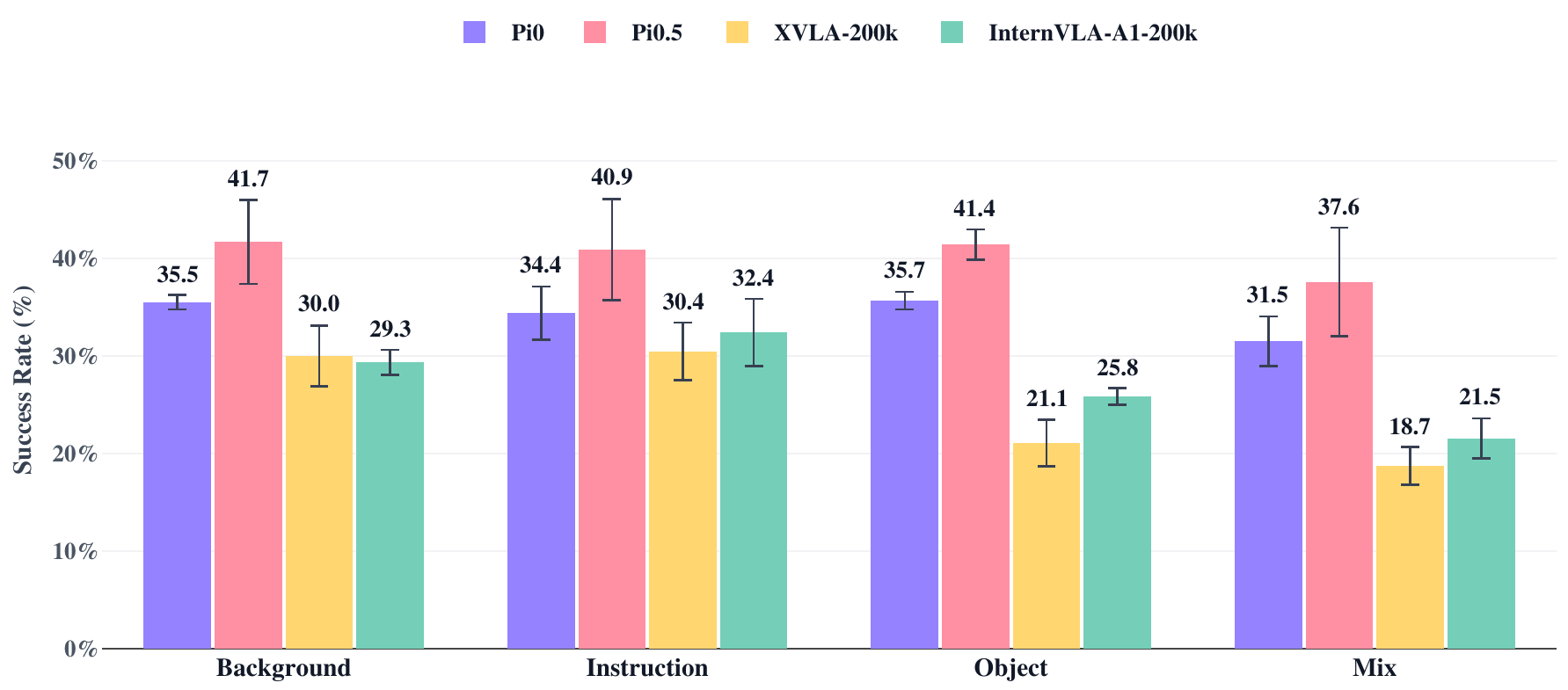}
    \caption{Test SR across four generalization dimensions.
}
\label{fig:generalization}
\end{wrapfigure}

\paragraph{Generalization across axes.}\label{subsec:gen_dims}
Figure~\ref{fig:generalization} decomposes test SR across four generalization dimensions: Background, Instruction, Object, and Mix, corresponding to unseen background, paraphrased instruction, unseen object instance, and their joint perturbation, respectively. Mix is the most challenging condition for every model, while the effect of Object-only shifts varies by model.
All four models maintain $29$--$42\%$ SR under Background and Instruction perturbations, suggesting that their perceptual and language grounding remain relatively robust when object physics is unchanged.
Object shifts yield $21$--$41\%$ SR, with their effect varying substantially by model.
The joint Mix setting further lowers SR to $19$--$38\%$, showing that compositional distribution shifts amplify failure modes beyond any single perturbation. Overall, $\pi_{0.5}$ is the most robust baseline, leading on Background, Instruction, Object, and Mix.


\section{Pretraining Sensitivity Across Benchmarks}
\label{app:pretrain}

A central question for evaluating generalist manipulation policies is: \textbf{Can the benchmark reveal the performance differences associated with large-scale pretraining?}
We address this question by comparing four representative architectures -- $\pi_0$~\citep{black2024pi_0}, $\pi_{0.5}$~\citep{intelligence2025pi_}, XVLA~\citep{zheng2025x}, and Fast-WAM~\citep{yuan2026fast} -- under two training regimes on three benchmarks: EBench, LIBERO~\citep{liu2023libero,fei2025libero}, and RoboTwin~2.0~\citep{chen2025robotwin}.
In the \textbf{pretrained} regime, we evaluate the released checkpoint fine-tuned on the benchmark's training split.
In the \textbf{from-scratch} regime, we initialize the same architecture randomly and train it only on the benchmark's training split. Fast-WAM has no released pretrained checkpoint and therefore appears only in the from-scratch regime.

Specifically, we additionally train $\pi_0$, $\pi_{0.5}$, XVLA, and Fast-WAM from scratch on EBench. LIBERO scores are average success rates over the four official suites Spatial, Object, Goal, and Long; RoboTwin~2.0 scores are average success rates over the hard-task split.
Pretrained $\pi_0$, $\pi_{0.5}$, and XVLA values on LIBERO are taken from the respective model release papers;
pretrained values on RoboTwin~2.0 Hard are taken from LingBot-VA~\citep{li2026causal} for $\pi_0$ and $\pi_{0.5}$ and from MOTUS~\citep{bi2026motus} for XVLA.
The from-scratch entries for $\pi_{0.5}$ and XVLA on LIBERO and RoboTwin~2.0 are reported as `--' because these architectures are not trained from scratch on these benchmarks in published work, and we do not run such ablations ourselves;
the Fast-WAM row supplies the without-pretrain data point on both benchmarks.

\begin{table}[h]
\centering
\caption{Pretraining comparison across EBench, LIBERO, and RoboTwin~2.0. The \emph{Pretrain} column indicates whether each row evaluates the released checkpoint or a model trained from random initialization on the benchmark's training split.
$^{\dagger}$\,The from-scratch $\pi_0$ entries on LIBERO and RoboTwin~2.0 Hard are taken from the StarVLA-$\pi$ configuration reported in~\citet{community2026starvla}, which follows the $\pi_0$ architecture with a Qwen3-VL-4B VLM backbone and is trained from random initialization on each benchmark's training split. *Results are cited from~\citep{ye2026starvla}. }
\label{tab:pretrain}
\small
\resizebox{\textwidth}{!}{%
\begin{tabular}{@{}lccccc@{}}
\toprule
\textbf{Model} & \textbf{Pretrain} & \textbf{EBench-\textsc{Test}(SR)} & \textbf{EBench-\textsc{Test}(Score)} & \textbf{LIBERO-Avg} & \textbf{RoboTwin~2.0-Hard} \\
\midrule
XVLA~\citep{zheng2025x}            & $\times$ & 15.7 & 27.7 & -- & -- \\
$\pi_0$~\citep{black2024pi_0}         & $\times$ &  24.8 & 39.2 & 95.7$^{\dagger}$ & 88.8$^{\dagger}$ \\
$\pi_{0.5}$~\citep{intelligence2025pi_}     & $\times$ &  15.9 &  25.0 & -- & -- \\
Fast-WAM~\citep{yuan2026fast}        & $\times$ &   25.0 &  36.6 & 97.6 & 91.8 \\
\midrule
XVLA~\citep{zheng2025x}            & \checkmark & 24.7 \,\textcolor{ForestGreen}{\textbf{(+9.0)}}  & 38.7 \,\textcolor{ForestGreen}{\textbf{(+11.0)}} & 98.1 & --  \\
$\pi_0$~\citep{black2024pi_0}         & \checkmark & 34.4 \,\textcolor{ForestGreen}{\textbf{(+9.6)}} & 50.9 \,\textcolor{ForestGreen}{\textbf{(+11.7)}} & 94.1$^{*}$ & 58.4$^{*}$ \\
$\pi_{0.5}$~\citep{intelligence2025pi_}     & \checkmark & 41.0 \,\textcolor{ForestGreen}{\textbf{(+25.1)}} & 56.8 \,\textcolor{ForestGreen}{\textbf{(+31.8)}} & 96.9$^{*}$ & 76.8$^{*}$ \\
\bottomrule
\end{tabular}%
}
\end{table}

\paragraph{Results.}
On EBench, pretraining consistently improves all three architectures, yielding substantial gains of 9.0--25.1 SR points: $\pi_0$ increases from 24.8\% to 34.4\%, $\pi_{0.5}$ from 15.9\% to 41.0\%, and XVLA from 15.7\% to 24.7\%.
In contrast, LIBERO exhibits near-saturated performance across all evaluated models, with scores ranging from 94.1\% to 98.1\%; notably, the from-scratch $\pi_0^\dagger$ variant achieves 95.7\%, slightly surpassing the pretrained $\pi_0$ result of 94.1\%.
RoboTwin~2.0 Hard shows a different trend: the two available from-scratch baselines outperform all reported pretrained baselines, achieving 88.8--91.8\% SR compared with 58.4--76.8\% for the pretrained counterparts. Near-ceiling performance on LIBERO leaves limited headroom for further improvement, whereas the reported cross-model comparisons on RoboTwin~2.0 Hard do not show an advantage for pretrained models.
EBench, in contrast, exhibits large and consistent within-architecture pretraining gains of 9.0--25.1 SR points, supporting its use for evaluating the contribution of large-scale pretraining to generalist policy performance.

\section{Limitations}
\label{sec:limitations}

EBench operates entirely in simulation, and we do not claim that simulation scores predict real-robot performance. However, we would like to treat EBench as a reproducible screening substrate that precedes physical evaluation rather than replacing it. We will also study the correlation between sim and real evaluation based on EBench tasks in future work. 
The 26-task suite covers 9 scene categories sparsely, so scene-level rankings should be read as preliminary; expanding toward $50$ or more tasks is on our roadmap and will unlock multi-way regression in place of permutation tests.


\section{Conclusion}
\label{sec:conclusion}

We presented EBench, a simulation benchmark for generalist embodied manipulation that places long-horizon, dexterous-and-precise, and mobile manipulation under a single evaluation protocol, a combination that current public benchmarks individually approximate but jointly avoid.
EBench pairs 26 capability-tagged tasks with four controlled generalization dimensions, drawn from a test asset pool that is disjoint from the training pool, and supplies them with a $91.4$ hours dataset synthesised through two complementary tracks: teleoperation for dexterous-and-precise tasks, and key-frame poses with cuRobo for mobile and long-horizon tasks.
Applied to $\pi_0$, $\pi_{0.5}$, XVLA, and InternVLA-A1, EBench shows that VLAs differ at the scalar-SR level and by tens of points along interpretable axes, namely operating mode, precision, horizon, and atomic skill, and follow distinct fit--generalise trajectories as training proceeds.
$\pi_{0.5}$ leads aggregate test performance, while InternVLA-A1, $\pi_0$, and XVLA each dominate disjoint subsets of the capability space, and the field has not converged on a single inductive bias.
Beyond capability profiling, a pretraining sensitivity analysis (\Cref{app:pretrain}) shows that, among EBench, LIBERO, and RoboTwin~2.0, 
EBench provides the clearest evidence of pretraining benefits: large-scale pretraining improves every architecture by 9.0--25.1 SR points on EBench, whereas the reported LIBERO and RoboTwin~2.0 comparisons show limited or even reversed gains for pretrained models. The tasks, synthesized dataset, evaluation code, and all splits are publicly released so that future generalist policies can be diagnosed along the same five capability axes.


\bibliography{ebench_corl2026}

@inproceedings{bi2026motus,
  title={Motus: A unified latent action world model},
  author={Bi, Hongzhe and Tan, Hengkai and Xie, Shenghao and Wang, Zeyuan and Huang, Shuhe and Liu, Haitian and Zhao, Ruowen and Feng, Yao and Xiang, Chendong and Rong, Yinze and others},
  booktitle={Proceedings of the IEEE/CVF Conference on Computer Vision and Pattern Recognition},
  pages={35101--35113},
  year={2026}
}

@article{team2024octo,
  title={Octo: An open-source generalist robot policy},
  author={Team, Octo Model and Ghosh, Dibya and Walke, Homer and Pertsch, Karl and Black, Kevin and Mees, Oier and Dasari, Sudeep and Hejna, Joey and Kreiman, Tobias and Xu, Charles and others},
  journal={arXiv preprint arXiv:2405.12213},
  year={2024}
}

@article{kim2024openvla,
  title={Openvla: An open-source vision-language-action model},
  author={Kim, Moo Jin and Pertsch, Karl and Karamcheti, Siddharth and Xiao, Ted and Balakrishna, Ashwin and Nair, Suraj and Rafailov, Rafael and Foster, Ethan and Lam, Grace and Sanketi, Pannag and others},
  journal={arXiv preprint arXiv:2406.09246},
  year={2024}
}

@article{black2024pi_0,
  title={{$\pi_0$: A Vision-Language-Action Flow Model for General Robot Control}},
  author={Black, Kevin and Brown, Noah and Driess, Danny and Esmail, Adnan and Equi, Michael and Finn, Chelsea and Fusai, Niccolo and Groom, Lachy and Hausman, Karol and Ichter, Brian and others},
  journal={arXiv preprint arXiv:2410.24164},
  year={2024}
}

@article{intelligence2025pi_,
  title={{$\pi_{0.5}$: A Vision-Language-Action Model with Open-World Generalization}},
  author={Intelligence, Physical and Black, Kevin and Brown, Noah and Darpinian, James and Dhabalia, Karan and Driess, Danny and Esmail, Adnan and Equi, Michael and Finn, Chelsea and Fusai, Niccolo and others},
  journal={arXiv preprint arXiv:2504.16054},
  year={2025}
}

@article{zheng2025x,
  title={X-vla: Soft-prompted transformer as scalable cross-embodiment vision-language-action model},
  author={Zheng, Jinliang and Li, Jianxiong and Wang, Zhihao and Liu, Dongxiu and Kang, Xirui and Feng, Yuchun and Zheng, Yinan and Zou, Jiayin and Chen, Yilun and Zeng, Jia and others},
  journal={arXiv preprint arXiv:2510.10274},
  year={2025}
}

@article{cai2026internvla,
  title={InternVLA-A1: Unifying Understanding, Generation and Action for Robotic Manipulation},
  author={Cai, Junhao and Cai, Zetao and Cao, Jiafei and Chen, Yilun and He, Zeyu and Jiang, Lei and Li, Hang and Li, Hengjie and Li, Yang and Liu, Yufei and others},
  journal={arXiv preprint arXiv:2601.02456},
  year={2026}
}

@article{yuan2026fast,
  title={Fast-WAM: Do World Action Models Need Test-time Future Imagination?},
  author={Yuan, Tianyuan and Dong, Zibin and Liu, Yicheng and Zhao, Hang},
  journal={arXiv preprint arXiv:2603.16666},
  year={2026}
}

@article{li2026causal,
  title={Causal World Modeling for Robot Control},
  author={Li, Lin and Zhang, Qihang and Luo, Yiming and Yang, Shuai and Wang, Ruilin and Han, Fei and Yu, Mingrui and Gao, Zelin and Xue, Nan and Zhu, Xing and others},
  journal={arXiv preprint arXiv:2601.21998},
  year={2026}
}

@article{mees2022calvin,
  title={Calvin: A benchmark for language-conditioned policy learning for long-horizon robot manipulation tasks},
  author={Mees, Oier and Hermann, Lukas and Rosete-Beas, Erick and Burgard, Wolfram},
  journal={IEEE Robotics and Automation Letters},
  volume={7},
  number={3},
  pages={7327--7334},
  year={2022},
  publisher={IEEE}
}

@article{mu2021maniskill,
  title={Maniskill: Generalizable manipulation skill benchmark with large-scale demonstrations},
  author={Mu, Tongzhou and Ling, Zhan and Xiang, Fanbo and Yang, Derek and Li, Xuanlin and Tao, Stone and Huang, Zhiao and Jia, Zhiwei and Su, Hao},
  journal={arXiv preprint arXiv:2107.14483},
  year={2021}
}

@inproceedings{o2024open,
  title={Open x-embodiment: Robotic learning datasets and rt-x models: Open x-embodiment collaboration 0},
  author={O’Neill, Abby and Rehman, Abdul and Maddukuri, Abhiram and Gupta, Abhishek and Padalkar, Abhishek and Lee, Abraham and Pooley, Acorn and Gupta, Agrim and Mandlekar, Ajay and Jain, Ajinkya and others},
  booktitle={2024 IEEE International Conference on Robotics and Automation (ICRA)},
  pages={6892--6903},
  year={2024},
  organization={IEEE}
}

@inproceedings{savva2019habitat,
  title={Habitat: A platform for embodied ai research},
  author={Savva, Manolis and Kadian, Abhishek and Maksymets, Oleksandr and Zhao, Yili and Wijmans, Erik and Jain, Bhavana and Straub, Julian and Liu, Jia and Koltun, Vladlen and Malik, Jitendra and others},
  booktitle={Proceedings of the IEEE/CVF international conference on computer vision},
  pages={9339--9347},
  year={2019}
}

@article{ye2026starvla,
  title={{StarVLA-$\alpha$: Reducing Complexity in Vision-Language-Action Systems}},
  author={Ye, Jinhui and Gao, Ning and Yang, Senqiao and Zheng, Jinliang and Wang, Zixuan and Chen, Yuxin and Chen, Pengguang and Chen, Yilun and Liu, Shu and Jia, Jiaya},
  journal={arXiv preprint arXiv:2604.11757},
  year={2026}
}

@article{community2026starvla,
  title={StarVLA: A Lego-like Codebase for Vision-Language-Action Model Developing},
  author={Community, StarVLA},
  journal={arXiv preprint arXiv:2604.05014},
  year={2026}
}

@article{fei2025libero,
  title={Libero-plus: In-depth robustness analysis of vision-language-action models},
  author={Fei, Senyu and Wang, Siyin and Shi, Junhao and Dai, Zihao and Cai, Jikun and Qian, Pengfang and Ji, Li and He, Xinzhe and Zhang, Shiduo and Fei, Zhaoye and others},
  journal={arXiv preprint arXiv:2510.13626},
  year={2025}
}

@article{bjorck2025gr00t,
  title={Gr00t n1: An open foundation model for generalist humanoid robots},
  author={Bjorck, Johan and Casta{\~n}eda, Fernando and Cherniadev, Nikita and Da, Xingye and Ding, Runyu and Fan, Linxi and Fang, Yu and Fox, Dieter and Hu, Fengyuan and Huang, Spencer and others},
  journal={arXiv preprint arXiv:2503.14734},
  year={2025}
}

@article{liu2023libero,
  title={Libero: Benchmarking knowledge transfer for lifelong robot learning},
  author={Liu, Bo and Zhu, Yifeng and Gao, Chongkai and Feng, Yihao and Liu, Qiang and Zhu, Yuke and Stone, Peter},
  journal={Advances in Neural Information Processing Systems},
  volume={36},
  pages={44776--44791},
  year={2023}
}

@inproceedings{li2023behavior,
  title={Behavior-1k: A benchmark for embodied ai with 1,000 everyday activities and realistic simulation},
  author={Li, Chengshu and Zhang, Ruohan and Wong, Josiah and Gokmen, Cem and Srivastava, Sanjana and Mart{\'\i}n-Mart{\'\i}n, Roberto and Wang, Chen and Levine, Gabrael and Lingelbach, Michael and Sun, Jiankai and others},
  booktitle={Conference on Robot Learning},
  pages={80--93},
  year={2023},
  organization={PMLR}
}

@article{nasiriany2024robocasa,
  title={Robocasa: Large-scale simulation of everyday tasks for generalist robots},
  author={Nasiriany, Soroush and Maddukuri, Abhiram and Zhang, Lance and Parikh, Adeet and Lo, Aaron and Joshi, Abhishek and Mandlekar, Ajay and Zhu, Yuke},
  journal={arXiv preprint arXiv:2406.02523},
  year={2024}
}

@article{wang2024grutopia,
  title={Grutopia: Dream general robots in a city at scale},
  author={Wang, Hanqing and Chen, Jiahe and Huang, Wensi and Ben, Qingwei and Wang, Tai and Mi, Boyu and Huang, Tao and Zhao, Siheng and Chen, Yilun and Yang, Sizhe and others},
  journal={arXiv preprint arXiv:2407.10943},
  year={2024}
}

@article{james2020rlbench,
  title={Rlbench: The robot learning benchmark \& learning environment},
  author={James, Stephen and Ma, Zicong and Arrojo, David Rovick and Davison, Andrew J},
  journal={IEEE Robotics and Automation Letters},
  volume={5},
  number={2},
  pages={3019--3026},
  year={2020},
  publisher={IEEE}
}

@inproceedings{deitke2023objaverse,
  title={Objaverse: A universe of annotated 3d objects},
  author={Deitke, Matt and Schwenk, Dustin and Salvador, Jordi and Weihs, Luca and Michel, Oscar and VanderBilt, Eli and Schmidt, Ludwig and Ehsani, Kiana and Kembhavi, Aniruddha and Farhadi, Ali},
  booktitle={Proceedings of the IEEE/CVF conference on computer vision and pattern recognition},
  pages={13142--13153},
  year={2023}
}

@inproceedings{xiang2020sapien,
  title={Sapien: A simulated part-based interactive environment},
  author={Xiang, Fanbo and Qin, Yuzhe and Mo, Kaichun and Xia, Yikuan and Zhu, Hao and Liu, Fangchen and Liu, Minghua and Jiang, Hanxiao and Yuan, Yifu and Wang, He and others},
  booktitle={Proceedings of the IEEE/CVF conference on computer vision and pattern recognition},
  pages={11097--11107},
  year={2020}
}

@inproceedings{liu2025rdt,
  title={Rdt-1b: a diffusion foundation model for bimanual manipulation},
  author={Liu, Songming and Wu, Lingxuan and Li, Bangguo and Tan, Hengkai and Chen, Huayu and Wang, Zhengyi and Xu, Ke and Su, Hang and Zhu, Jun},
  booktitle={International Conference on Learning Representations},
  volume={2025},
  pages={29982--30009},
  year={2025}
}

@article{chi2025diffusion,
  title={Diffusion policy: Visuomotor policy learning via action diffusion},
  author={Chi, Cheng and Xu, Zhenjia and Feng, Siyuan and Cousineau, Eric and Du, Yilun and Burchfiel, Benjamin and Tedrake, Russ and Song, Shuran},
  journal={The International Journal of Robotics Research},
  volume={44},
  number={10-11},
  pages={1684--1704},
  year={2025},
  publisher={Sage Publications Sage UK: London, England}
}

@inproceedings{mu2025robotwin,
  title={Robotwin: Dual-arm robot benchmark with generative digital twins},
  author={Mu, Yao and Chen, Tianxing and Chen, Zanxin and Peng, Shijia and Lan, Zhiqian and Gao, Zeyu and Liang, Zhixuan and Yu, Qiaojun and Zou, Yude and Xu, Mingkun and others},
  booktitle={Proceedings of the computer vision and pattern recognition conference},
  pages={27649--27660},
  year={2025}
}

@inproceedings{gao2025genmanip,
  title={Genmanip: Llm-driven simulation for generalizable instruction-following manipulation},
  author={Gao, Ning and Chen, Yilun and Yang, Shuai and Chen, Xinyi and Tian, Yang and Li, Hao and Huang, Haifeng and Wang, Hanqing and Wang, Tai and Pang, Jiangmiao},
  booktitle={Proceedings of the Computer Vision and Pattern Recognition Conference},
  pages={12187--12198},
  year={2025}
}

@article{chen2025robotwin,
  title={Robotwin 2.0: A scalable data generator and benchmark with strong domain randomization for robust bimanual robotic manipulation},
  author={Chen, Tianxing and Chen, Zanxin and Chen, Baijun and Cai, Zijian and Liu, Yibin and Li, Zixuan and Liang, Qiwei and Lin, Xianliang and Ge, Yiheng and Gu, Zhenyu and others},
  journal={arXiv preprint arXiv:2506.18088},
  year={2025}
}

@article{chen2026rmbench,
  title={Rmbench: Memory-dependent robotic manipulation benchmark with insights into policy design},
  author={Chen, Tianxing and Wang, Yuran and Li, Mingleyang and Qin, Yan and Shi, Hao and Li, Zixuan and Hu, Yifan and Zhang, Yingsheng and Wang, Kaixuan and Chen, Yue and others},
  journal={arXiv preprint arXiv:2603.01229},
  year={2026}
}

@inproceedings{sundaralingam2023curobo,
  title={Curobo: Parallelized collision-free robot motion generation},
  author={Sundaralingam, Balakumar and Hari, Siva Kumar Sastry and Fishman, Adam and Garrett, Caelan and Van Wyk, Karl and Blukis, Valts and Millane, Alexander and Oleynikova, Helen and Handa, Ankur and Ramos, Fabio and others},
  booktitle={2023 IEEE International Conference on Robotics and Automation (ICRA)},
  pages={8112--8119},
  year={2023},
  organization={IEEE}
}

@article{zhong2026internscenes,
  title={Internscenes: A large-scale simulatable indoor scene dataset with realistic layouts},
  author={Zhong, Weipeng and Cao, Peizhou and Jin, Yichen and Li, Luo and Cai, Wenzhe and Lin, Jingli and Wang, Hanqing and Lyu, Zhaoyang and Wang, Tai and XU, Xudong and others},
  journal={Advances in Neural Information Processing Systems},
  volume={38},
  year={2026}
}

@article{li2024evaluating,
  title={Evaluating real-world robot manipulation policies in simulation},
  author={Li, Xuanlin and Hsu, Kyle and Gu, Jiayuan and Pertsch, Karl and Mees, Oier and Walke, Homer Rich and Fu, Chuyuan and Lunawat, Ishikaa and Sieh, Isabel and Kirmani, Sean and others},
  journal={arXiv preprint arXiv:2405.05941},
  year={2024}
}

\appendix
\appendix


\section{Implementation Details}
\label{app:implementation}

\paragraph{Baselines.}
To make cross-model comparisons fair, every baseline in this paper is trained and rolled out under a common training budget and action-chunk schedule, regardless of architecture.
We use a global batch size of $128$, a relative action-chunk prediction horizon of $50$ timesteps, and an open-loop application horizon of $30$ steps -- the policy predicts a $50$-step chunk at every replanning step, but only the first $30$ actions are executed in the environment before the next prediction is issued.
The resulting $20$-step lookahead buffer absorbs the model's inference latency without introducing closed-loop instability.
All other architecture-specific hyperparameters (optimizer, learning-rate schedule, dropout, action normalisation, etc.) follow the official open-source repository of each model unchanged.

\paragraph{Metrics.}
Each episode reports a binary success signal, SR, and a continuous task score that rewards partial progress such as correctly placing 2 of 3 objects.
We report both metrics because the score captures ``near misses'' that binary success discards, particularly on long-horizon tasks where individual sub-goals may be partially satisfied.

\paragraph{Camera configurations.}
EBench supports two primary camera configurations for studying visual perspective effects:
\textbf{Headview:} An egocentric camera mounted on the robot head, providing a local, first-person perspective aligned with the end-effector gaze.
\textbf{Overview:} A bird's-eye camera positioned above the scene, providing a global, top-down perspective of the workspace and surrounding environment.
Both configurations include left and right auxiliary cameras for stereo cues.

\paragraph{Computational cost.}
A full validation run takes approximately 30 minutes on eight RTX~4090 GPUs using the distributed evaluation toolkit.
The complete benchmark, validation plus test, completes in under two hours on the same hardware, enabling rapid iterative development.

\section{Task-Level Analysis}
\label{app:capability_summary}

\subsection{Task-Level Complementarity and Hard Tasks}
\label{app:task_level}

\begin{figure}[h]
\centering
\includegraphics[width=\textwidth]{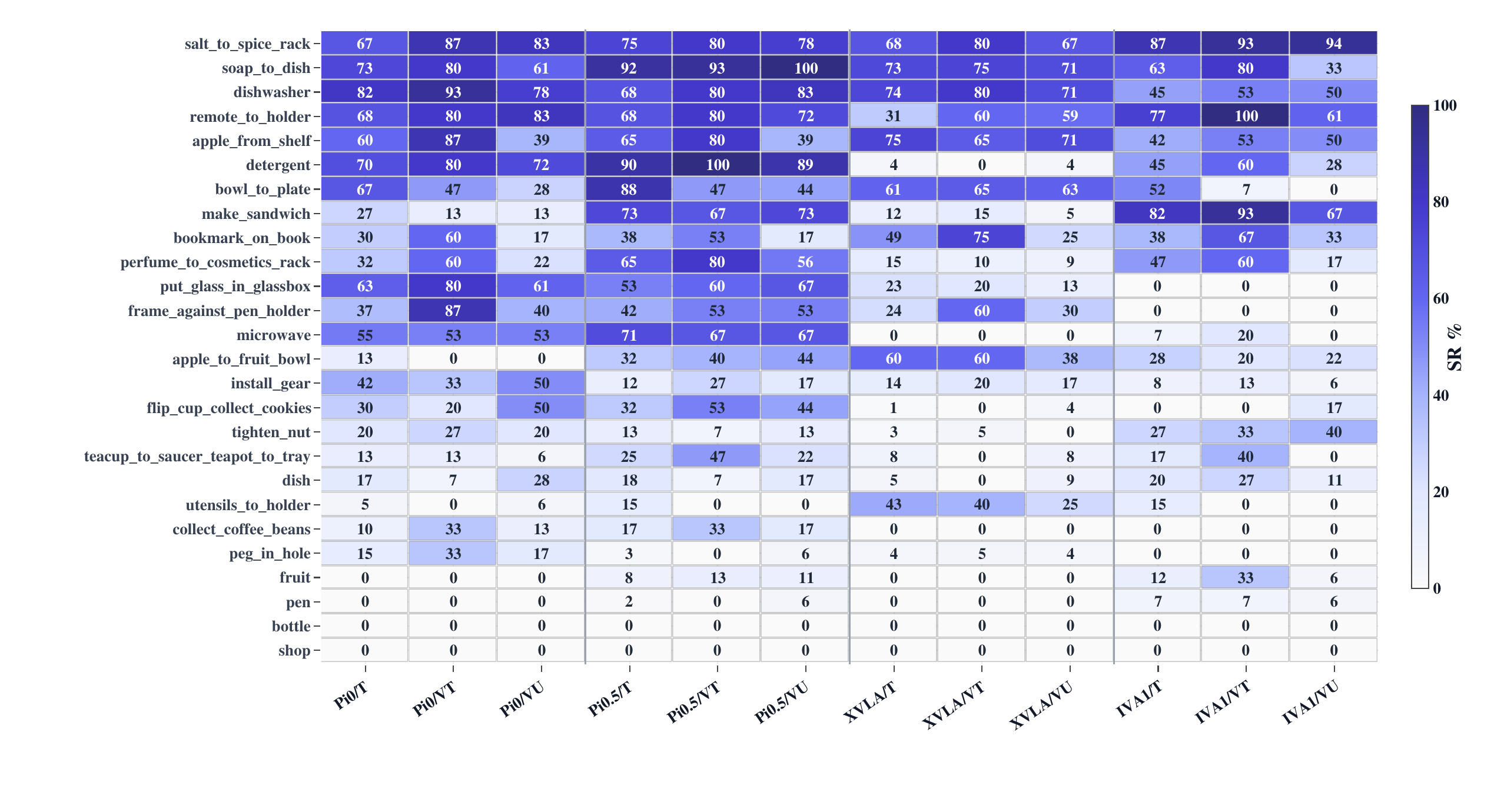}
\caption{Per-task success-rate heatmap across the four baselines and three splits. VT: \textsc{Validation-Train}, VU: \textsc{Validation-Unseen}, T: \textsc{Test}.}
\label{fig:per_task_heatmap}
\end{figure}

Figure~\ref{fig:per_task_heatmap} renders the full per-task picture: rows are the 26 tasks (sorted top-to-bottom by the sum across all model--split cells), columns are the four baselines crossed with the three splits (\textsc{Validation-Train}, \textsc{Validation-Unseen}, \textsc{Test}), and the color encodes per-task success rate.
Two structural phenomena pop out immediately from this view.

\textbf{Per-task complementarity.}
The $\pi$-family $\pi_0$ and $\pi_{0.5}$, taken together, and XVLA exhibit strong complementarity that the cluster bars in the main paper smooth over.
Tasks where the $\pi$-family outperforms XVLA by the largest Test-SR margin include \texttt{detergent} at $+86\%$, \texttt{microwave} at $+71\%$, and \texttt{make\_sandwich} at $+61\%$; these read as a vertical pair of dark cells in the $\pi$-family blocks above pale cells in the XVLA block.
Conversely, XVLA wins on \texttt{apple\_to\_fruit\_bowl} at $+28\%$, \texttt{utensils\_to\_holder} at $+28\%$, and \texttt{bookmark\_on\_book} at $+11\%$, which produce the inverse stripe -- pale $\pi$-family cells against a dark XVLA column.
InternVLA-A1 sits between the two families on most rows but resolves several of the $\pi$-family weaknesses (notably \texttt{make\_sandwich} and \texttt{salt\_to\_spice\_rack}) at the cost of being weaker on the dexterous tabletop tail.

\textbf{Universally-hard tasks.}
At the bottom of the heatmap, only two rows stay at zero across every column and every split: \texttt{shop} and \texttt{bottle}.
All four baselines score $0\%$ SR on these tasks across all three splits, indicating that they lie beyond the current frontier of generalist policy capability. We therefore propose these two tasks as a small ``hard suite'' for future generalist policies: even modest non-zero success on the suite would indicate a meaningful progress, whereas aggregate scores over the full 26-task benchmark can be dominated by the easier majority.


\end{document}